\documentclass{article}
\usepackage{spconf,amsmath,graphicx}

\usepackage{booktabs}
\usepackage{multirow}

\usepackage{url}
 
\title{Generating human readable transcript \\ for automatic speech recognition with pre-trained language model}
\name{Junwei Liao$^{1*}$\thanks{Work done as an intern at Microsoft}, Yu Shi$^{2}$, Ming Gong$^3$, Linjun Shou$^3$, Sefik Eskimez$^2$, Liyang Lu$^2$, Hong Qu$^{1}$, Michael Zeng$^2$}
\address{$^1$University of Electronic Science and Technology of China \\ $^2$Microsoft Cognitive Services Research Group, USA \\ $^3$Microsoft STCA NLP Group \\
\texttt{\normalsize junwei.liao@outlook.com, hongqu@uestc.edu.cn,} \\
\texttt{\normalsize \{yushi, migon, lisho, seeskime, liyang.lu, nzeng\}@microsoft.com}
}
\copyrightnotice{\begin{tabular}[t]{@{}l@{}}© 2021 IEEE. Personal use of this material is permitted. Permission from IEEE must be obtained for all other uses, in any current or\\future media, including reprinting/republishing this material for advertising or promotional purposes, creating new collective works,\\for resale or redistribution to servers or lists, or reuse of any copyrighted component of this work in other works.\end{tabular}}

\begin{document}
%
\maketitle
\begin{abstract}
Modern Automatic Speech Recognition (ASR) systems can achieve high performance in terms of recognition accuracy. However, a perfectly accurate transcript still can be challenging to read due to disfluency, filter words, and other errata common in spoken communication. Many downstream tasks and human readers rely on the output of the ASR system; therefore, errors introduced by the speaker and ASR system alike will be propagated to the next task in the pipeline.
In this work, we propose an ASR post-processing model that aims to transform the incorrect and noisy ASR output into a readable text for humans and downstream tasks. We leverage the Metadata Extraction (MDE) corpus to construct a task-specific dataset for our study. Since the dataset is small, we propose a novel data augmentation method and use a two-stage training strategy to fine-tune the RoBERTa pre-trained model.
On the constructed test set, our model outperforms a production two-step pipeline-based post-processing method by a large margin of 13.26 on readability-aware WER (RA-WER) and 17.53 on BLEU metrics.
Human evaluation also demonstrates that our method can generate more human-readable transcripts than the baseline method.

\end{abstract}
\section{Introduction}
\label{sec:intro}

With the rapid development of speech-to-text technologies, ASR systems have achieved high recognition accuracy, even beating the performance of professional human transcribers on conversational telephone speech in terms of Word Error Rate (WER) \cite{xiong2018microsoft}.
ASR systems bring convenience to users in many scenarios. However, colloquial speech is fraught with disfluency, informal words, and other noises that make it difficult to understand. While ASR systems do a great job in recognizing which words are said, its verbatim transcription creates many problems for modern applications that must comprehend the meaning and intent of what is said. 
The existence of the defects in speech transcription will significantly harm the experience of the application users if the system cannot handle them well.

There is a long line of previous works focusing on making ASR transcripts more human-readable, which is referred to as metadata extraction (MDE) \cite{liu2005structural}. MDE breaks down the goal into several classification tasks on top of verbatim transcription. Most of the MDE systems use both textual and prosodic information combined by Hidden Markov Model \cite{liu2004icsi, tomalin2004rt04}, Maximum Entropy or Conditional Random Fields methods \cite{liu2004icsi,liu2004comparing}. While MDE improves the readability of ASR transcripts to a certain extent, it ignores the recognition errors introduced by ASR system. Thus MDE needs to work with other ASR post-processing components such as language model rescoring to provide the final human-readable transcript.

In this paper, We propose an end-to-end ASR post-processing model for readability (APR). Readability in this context refers to having proper segmentation, capitalization, fluency, and without any error, so our model aims to transform the ASR output into an error-free and readable text in one shot.
Our model is based on RoBERTa \cite{liu2019roberta}, which is a pre-trained language model used for NLU tasks. Inspired by UniLM \cite{dong2019unified}, which applies self-attention masks on BERT \cite{devlin2019bert} to convert it into a sequence-to-sequence model, we adapt RoBERTa towards generative for NLG task.

Since there is no off-the-shelf dataset for the proposed task, we construct the desired dataset from the RT-03 MDE Training Data \cite{strassel2004rt}, which includes speech audio, human transcript, and annotation. 
We evaluate the proposed approaches on a test split of the constructed dataset and compare it with a production pipeline-based baseline.
Our model significantly outperforms baseline method 13.26 on RA-WER and 17.53 on BLEU metrics. Human evaluation also shows that the proposed model generates more human-readable transcripts for ASR output.

To solve the training data scarce problem, in addition to directly fine-tune the pre-trained language model, we propose a novel data augmentation method that synthesizes large-scale training data using Grammatical Error Correction (GEC) corpora followed by text-to-speech (TTS) and ASR. 
We adopt a two-stage training strategy, namely pre-training and fine-tuning, to better exploit the augmented data. This two-phase training not only learns useful information from the augmented data, but also avoids the risk of being overwhelmed and adversely affected by it.

\section{Model}
\label{sec:Model}

\begin{figure}[t]
\begin{minipage}[b]{1.0\linewidth}
  \centering
  \centerline{\includegraphics[width=8.5cm]{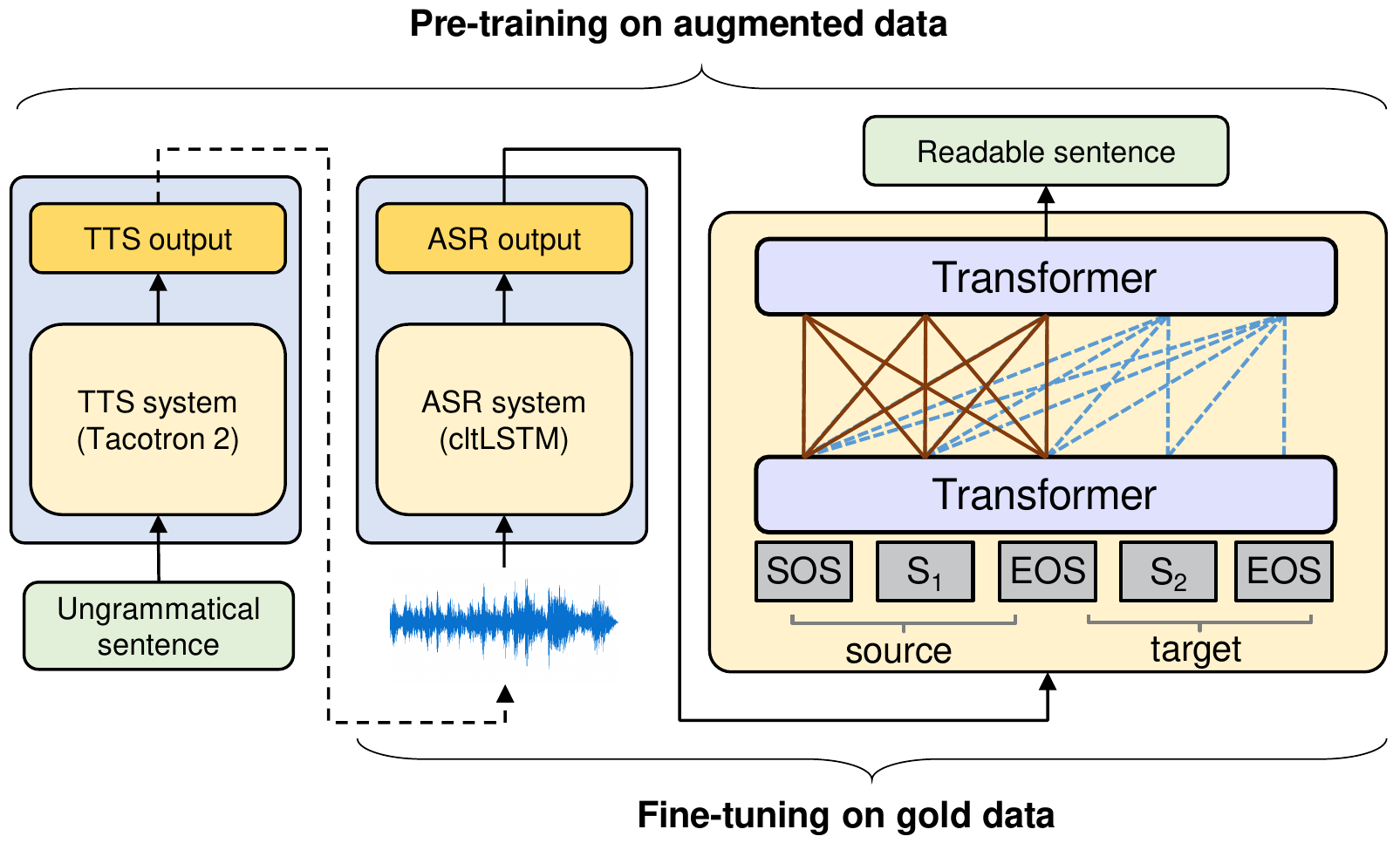}}
\end{minipage}
\caption{ASR post-processing model for readability based on modified ReBERTa architecture and data augmentation.}
\label{fig:model}
\vspace{-0.5cm}
\end{figure}

Figure \ref{fig:model} shows the whole picture of the ASR post-processing for readability.

\subsection{Speech recognition model}

The middle part in Figure~\ref{fig:model} is the ASR model. We use a contextual layer trajectory LSTM (cltLSTM) \cite{li2019improving} as our ASR model. The cltLSTM decouples the temporal modeling and target classification tasks with time and depth LSTMs respectively and incorporates future context frames to get more information for accurate acoustic modeling. The input feature is an 80-dimension log Mel filter bank for every 20 milliseconds (ms) of speech by using frame skipping \cite{miao2016simplifying}. The softmax layer has 9404 nodes to model the senone labels. Runtime decoding is performed using a 5-gram LM with a decoding graph of around 5 gigabytes (Gbs). The cltLSTM has 24-frame lookahead, which corresponds to a 480ms duration. 


\subsection{ASR post-processing model for readability}

The proposed APR model (the rightmost part in Figure~\ref{fig:model}) is based on RoBERTa \cite{liu2019roberta}, which is a robustly optimized BERT \cite{devlin2019bert} pre-training approach. Both BERT and RoBERTa have a single Transformer stack and are pre-trained only using bidirectional prediction, which makes them more discriminative than generative. However, \cite{hrinchuk2020correction} demonstrated the effectiveness of transfer learning from BERT to sequence-to-sequence task by initializing both encoder and decoder with pre-trained BERT in their speech recognition correction work. Inspired by this work and UniLM \cite{dong2019unified}, we apply self-attention masks on the RoBERTa model to convert it into a sequence-to-sequence generation model. To achieve whole-sentence prediction rather than only masked-position prediction, we use an autoregressive approach during the fine-tuning. 


\section{Data Augmentation}

Transformer-based models are usually trained on millions of parallel sentences and tend to easily overfit if the data is scarce. \cite{hrinchuk2020correction,liao2020improving} proposed two self-complementary regularization techniques to solve this problem. Besides initializing the model weights using the pre-trained language model mentioned in the previous section, the other solution is data augmentation. Following their cue, we propose a novel data augmentation method for APR. 

We synthesize large-scale training data using a grammatical error correction (GEC) dataset as the seed data. GEC aims to correct different kinds of errors such as spelling, punctuation, grammatical, and word choice errors
. The purpose of using it is to introduce more types of errors that are not specific to our ASR system and therefore make the APR model more general. The GEC dataset comes from restricted tracks of BEA 2019 shared task \cite{Bryant2019TheBS} and is composed of pairs of grammatically incorrect sentences and corresponding sentences corrected by a human annotator. 

First, we use a text-to-speech (TTS) system (the leftmost part in Figure~\ref{fig:model}) to convert the ungrammatical sentences to speech. Specifically, we use a Tacotron2 model \cite{shen2018natural}, which is composed of a recurrent sequence-to-sequence feature prediction network that maps character embeddings to mel-scale spectrograms, followed by a modified WaveNet model acting as a vocoder to synthesize time-domain waveforms from those spectrograms. We fed the grammatically incorrect sentences from the seed corpus into Tacotron2 to produce the audio files simulating human speakers. 

Then, these audio files are fed into our ASR system that outputs the corresponding transcripts. The resulting text contains both the grammatical errors found initially in the GEC dataset and the TTS+ASR pipeline errors. 
Finally, we pair the outputs of the ASR system and the original grammatical sentences as the training samples.
After the above process, we obtain about 1.1M sentence pairs.

\section{Experiments}
\label{sec:experiments}

\subsection{Gold dataset construction}
To the best of our knowledge, there is no off-the-shelf dataset that can be directly used in our APR task. Therefore we construct the desired dataset from MDE\footnote{https://catalog.ldc.upenn.edu/LDC2004T12} corpus.

We use the English Conversational Telephone Speech (CTS) in the released data,
of which the transcripts and annotations cover approximately over 40 hours of CTS audios of casual and conversational speech.
By parsing the annotation files, we get the transcript with metadata annotation, which, for example, uses
`\verb|/.|' for statement boundaries (SU), `\verb|<>|' for fillers, `\verb|[]|' for disfluency edit words, and `\verb|*|' for interruption points inside edit disfluencies.
The following example shows an ASR transcript with metadata annotation: 
\begin{verbatim}
and < uh > < you know > wash your clothes 
wherever you are /. and [ you ] * you 
really get used to the outdoors ./
\end{verbatim}
We generate a readable target transcript by cleaning up the metadata annotations, i.e. the deletable portion of edit disfluencies and fillers are removed, and each SU is presented as a separate line within the transcript. We also capitalize the first word of sentences to further improve the transcript's readability. Through the above steps, the transcript with metadata annotation becomes a readable text: \textit{And wash your clothes wherever you are. And you really get used to the outdoors.}

After the above processing, We obtain 27,355 readable transcripts in total. 
About 1K samples are extracted for validation and testing, respectively. We ensure that samples of training, validation, and testing come from different conversations. 
Audios are transcribed using the aforementioned ASR model, and the outputs are paired with readable targets.

\subsection{Baseline}

We use the production two-step post-processing pipeline of our ASR system as the baseline, namely n-best LM rescoring followed by inverse text normalization (ITN). This pipeline works well for sequentially improving speech recognition accuracy and display format for readability.
The language model is a stacked RNN with two unidirectional LSTM layers \cite{sundermeyer2012lstm}. 
ITN is configured to turn on capitalization, punctuation, and correcting simple grammatical errors for MDE test data. 
Since this pipeline lacks handling disfluencies in spoken language, for a fair comparison, we add a simple step to remove some disfluencies. Specifically, we remove the commonly used filled pauses (e.g., uh, um) and discourse markers (e.g., you know, I mean). We filter the repeated words and keep only one of them (e.g., I’m I’m$\rightarrow$I’m, it’s it’s$\rightarrow$it’s). 

We take advantage of the BLEU \cite{papineni2002bleu} score to measure the performance of the APR model. 
We also extend the conventional WER in speech recognition to readability-aware WER (RA-WER) by removing
the text normalization before calculating Levenshtein distance.

\subsection{Model Training}

As aforementioned, the augmented GEC data is valuable in generalizing the APR model. However, the source transcript does not have many disfluencies and other errata that often happen in spoken communication.
As a result, if we mix the augmented data with the gold data during model training, the massive augmented data tends to overwhelm the gold data and introduce unnecessary and even erroneous editing knowledge, which is undesirable for readability. To solve this problem, we follow \cite{zhang2019sequence}'s strategy. Specifically, we train the model using augmented data and gold data in two phases: pre-training and fine-tuning, respectively. 

We explore the effect of data augmentation with two configurations in our experiment. In the first configuration denoted as APR (FT), we directly fine-tune the proposed model initialized using the weight of pre-trained RoBERTa on constructed data. In the second configuration denoted as APR (PT + FT), we first pre-train model initialized with the weight of pre-trained RoBERTa on augmented data, then fine-tune model on constructed data. 

Our model is implemented using PyTorch on top of the Huggingface Transformers library\footnote{https://github.com/huggingface/Transformers} and based on RoBERTa-large architecture.
We train our model with Adam optimizer (lr=2.5e-6, $\beta_1$=0.9, $\beta_2$=0.999) for 3 epochs with linear warmup over the one-tenth of total steps and linear decay on a batch size of 8 for each GPU. We adopt the 50K byte-level BPE vocabulary used in RoBERTa so we can straight transfer its pre-trained weights. 
Each model is trained using 4 NVIDIA V100 GPUs with 32GB of memory and mixed precision.
We use the same training setup for all training stages.
In the fine-tuning stage, checkpoints are selected on the validation set, and the beam size for beam search is 5.


\section{Results}
\label{sec:results}

\subsection{Results analysis}

\begin{table}[t]
  \centering
    \resizebox{0.7\linewidth}{!}{
    \setlength{\tabcolsep}{7mm}{
    \begin{tabular}{lcc}
    \toprule
    \multicolumn{1}{c}{\textbf{Model}} & \textbf{RA-WER} & \textbf{BLEU} \\
    \midrule
    ASR transcript & 45.13 & 45.10 \\
    \quad + LM rescoring & 44.71 & 45.68 \\
    \quad + ITN & 38.77 & 53.39 \\
    \quad + RM disfluencies & \textbf{33.52} & \textbf{56.76} \\
    \midrule
    APR (FT) & 21.33 & 72.77 \\
    APR (PT + FT) & \textbf{20.26} & \textbf{74.29} \\
    \bottomrule
    \end{tabular}
    }
    }%
  \caption{Performance of APR models and baseline methods on the test set of gold data.}
  \label{tab:results}%
  \vspace{-0.5cm}
\end{table}%

\begin{table*}[t]
  \centering
    \resizebox{1\linewidth}{!}{
    \begin{tabular}{p{18.5em}p{18.5em}p{18.5em}p{18.5em}}
    \toprule
    \multicolumn{1}{c}{\textbf{ASR transcript}} & \multicolumn{1}{c}{\textbf{Ground truth}} & \multicolumn{1}{c}{\textbf{ASR + LM Rescoring + ITN}} & \multicolumn{1}{c}{\textbf{APR (PT + FT)}} \\
    \midrule
    yeah i don't believe they have to pay any uh like federal tax & I don't believe , they have to pay any federal tax . & Yeah, I don't believe they have to pay any. Uh, like federal tax, uh? & I don't believe, they have to pay any federal tax. \\
    \midrule
    yeah i i buy him every once in awhile and an i bought one and it was you know blah & I buy them every once in a while . And I bought one . And it was blah . & Yeah I I buy 'em every once \textbf{in awhile} and an I bought one and it was, you know blah. & I buy them every once \textbf{in a while}. And I bought one. And it was blah. \\
    \midrule
    they have as far as i'm concerned because i'm i'm not a big vegetable eater they have too many a yellow vegetables on the same day & They have too many yellow vegetables on the same day . & They have, as far as I'm concerned, because I'm I'm not a big vegetable eater. They have too many a yellow vegetables on the same day. & They have too many yellow vegetables on the same day. \\
    \midrule
    i just see a lot of a social and cultural differences that could a post problems with a puerto rico becoming a state & I just see a lot of social and cultural differences that could pose problems with Puerto Rico becoming a state . & I just see a lot of, uh, social and cultural differences that could \textbf{a post problems} with Puerto Rico becoming a state. & I just see a lot of social and cultural differences that could \textbf{cause problems} with Puerto Rico becoming a state. \\
    \bottomrule
    \end{tabular}
    }%
  \caption{Comparison of generated readable transcript using baseline method and proposed model. The bold parts of sentences are the correction of recognition errors.}
  \label{tab:case_analysis}%
  \vspace{-0.5cm}
\end{table*}%

In Table~\ref{tab:results}, we compare model performance on constructed test data. From the table, we can see that APR (PT + FT) model outperforms the baseline (ASR + LM rescoring + ITN + RM disfluencies) by a large margin of 13.26 RA-WER and 17.53 BLEU (absolute value). 

To better understand the readability contribution of each stage in the baseline method, we compute the metric scores between each stage's output and the reference sentence and put the results in the first group of Table~\ref{tab:results}. LM rescoring only brings 0.42 RA-WER reduction and 0.58 BLEU promotion. One explanation is that for highly casual and conversational speech, recognition correctness especially verbatim recognition only contributes a small part of the readability.
ITN further improves the scores by 5.94$\downarrow$ RA-WER and 7.71$\uparrow$ BLEU, but not that far. It's obvious that having proper capitalization and punctuation is helpful for readability but not enough. 
With a simple step of removing some disfluencies, the performance gain is significant (RA-WER 5.25, BLEU 3.37), which implies that disfluencies are the major factor for the gap between baseline and our proposed method.

The second group in Table~\ref{tab:results} shows the performance of the proposed model using different training strategies. The two-stage training (PT + FT) further reduces RA-WER by 1.07 point and increases BLEU by 1.52 point. The result proves the effectiveness of the proposed data augmentation method and the two-stage training strategy for the APR task.

\subsection{Case Analysis}
To conduct a qualitative analysis of readable transcript produced by the proposed model, we compare the output examples of our APR (PT + FT) model with the baseline method (ASR + LM rescoring + ITN).
In Table~\ref{tab:case_analysis}, we can see that both the baseline method and our model improve the readability of the ASR transcript by adding punctuation and capitalizing the names and the first word of sentences. But the baseline is verbatim by keeping all the words in the ASR transcript, which makes the sentence disfluent and leads to incorrect segmentation and punctuation. For instance, in the first example, the baseline wrongly segments the sentence and add a question mark instead of period
due to the influence of filler words (``uh'', ``like''). In contrast to the baseline, our model removes all words that cause the disfluency and adds correct punctuation in the appropriate place in the transcript to make it more readable. The third example better illustrates the capability of the proposed model on removing disfluencies. Our model removes all the repetitions, asides, and parentheticals to get the clear sentence same with the ground truth. 

Besides the punctuation, capitalization, and removing disfluencies, our model also corrects some recognition errors while the baseline method fails to do. We use bold type font to highlight the corrected errors in Table~\ref{tab:case_analysis}. An interesting finding is that the last example gets ``a post problems'' fixed to ``cause problems'' which is different than the ground truth ``pose  problems'' because the latter is not often used. It's arguable that although the original user input is ``pose'', our model's result is more readable for most human readers and machine applications if we don't consider personalization.

\subsection{Human evaluation}

Readability is subjective and the BLEU score and RA-WER may not be consistent with human perception. Thus, we also conduct a human evaluation on the Switchboard corpus \cite{godfrey1997switchboard}. 
Specifically, we conduct an A/B test to compare our model with the baseline method. To build a test set for human evaluation, we randomly chose 100 audio samples with source sentences length between 20 and 60 words. These audio samples are passed through the ASR system to get transcripts. We then use the baseline method and our model to generate the output text, respectively. 
Three annotators are shown the generated texts in random order and are asked to choose the more readable one. 
Each case receives three labels from three annotators. The final decision is made by a majority vote. 
Based on the above experiment design, the human evaluation result shows that annotators vote for the outputs of our model 70 times out of 100 cases (win rate 70\%), which means that our model was rated more readable than the baseline method. The two-sided binomial test on the result confirms that our model is statistically significantly more readable than the baseline method, with a p-value of less than 0.01.


\section{CONCLUSION}
\label{sec:conclusion}

In this work, we propose an ASR post-processing model for readability which is based on a modified RoBERTa pre-trained language model. Fine-tuned on our constructed data, the proposed model is capable of ``translating'' the ASR output to an error-free and readable transcript for human understanding and downstream tasks. Case analysis and human evaluation demonstrate that our model outperforms the traditional pipeline-based baseline method and generates a more readable transcript. 


\vfill\pagebreak



\bibliographystyle{IEEEbib}
\bibliography{strings,refs}

\end{document}